\documentclass[runningheads]{llncs}
\usepackage[T1]{fontenc}

\usepackage{amsmath}
\usepackage{amssymb}
\usepackage{subcaption}
\usepackage{diagbox}
\usepackage{url} 
\usepackage{graphicx} % Required for inserting images
\usepackage[ruled,vlined]{algorithm2e}

\usepackage{comment}
\usepackage{color}
\begin{document}
\title{EMASAM: a Computationally Efficient \\Sharpness-Aware Minimization \\via EMA-Guided Perturbations}
\titlerunning{EMASAM: a Computationally Efficient Sharpness-Aware Minimization}

\author{Tanapat Ratchatorn \and Masayuki Tanaka}

\authorrunning{T. Ratchatorn \and M. Tanaka}

\institute{Institute of Science Tokyo, Meguro, Tokyo, Japan \email{tratchatorn@vip.sc.eng.isct.ac.jp}}

\maketitle 

\begin{abstract}
Recent progress in optimization research has highlighted the sharpness of the loss landscape as a key factor in narrowing the generalization gap. Motivated by this insight, Sharpness-Aware Minimization (SAM) was proposed as a training strategy that enhances generalization. Despite the promising performance, SAM suffers from its twice computational cost due to its core algorithm requiring an extra gradient computation during the perturbation step. To overcome this limitation, we introduce Exponential Moving Average Sharpness-Aware Minimization (EMASAM), a computationally efficient variant of SAM. EMASAM does not require the loss gradient in the perturbation step. Instead, EMASAM defines the perturbation direction based on the discrepancy between the main model and the EMA shadow model. This perturbation travels away from the stable average position toward the less stable area, acting as a softer yet cheaper alternative to SAM's worst-case scenario perturbation. Moreover, since EMASAM's perturbation does not rely on noisy mini-batch gradients, it mitigates the gradient-induced instability inherent in SAM. Hence, EMASAM eliminates the need for an extra backpropagation while also preserving the generalization ability of the SAM-style training. Several experiments have been performed and confirm the efficiency and robustness of our method. The reproduction code is available at \url{https://www.vip.sc.eng.isct.ac.jp/proj/EMASAM}.

\keywords{Model Generalization \and Sharpness-Aware Minimization \and Exponential Moving Average \and Weight Perturbation \and Deep Learning}
\end{abstract}

\section{Introduction}
\label{sec:intro}
Recent progress in machine learning has led to the development of highly overparameterized models. While such models are excellent at memorizing training data, a significant challenge arises in their performance on unseen samples. This phenomenon, referred to as overfitting, results in a gap in performance between training and testing datasets~\cite{zhang2021understanding}. Therefore, improving model generalization has become a critical objective, as it ensures that models can perform reliably not only on the data used for training but also on previously unseen inputs.

To address the challenge of generalization, numerous studies have explored this problem from different perspectives. Some researchers have analyzed it through a Bayesian perspective to provide insights into generalization~\cite{mcallester1999pac,neyshabur2017pac}, while others have approached it using information theory~\cite{liang2019fisher}. Additional research efforts have examined how training hyperparameters such as the learning rate~\cite{li2019towards,chaudhari2018stochastic,DBLP:journals/corr/GoyalDGNWKTJH17} and batch size~\cite{keskar2016large} influence a model’s generalization ability. Various techniques have been proposed to improve model generalization. For example, Entropy-SGD employs local entropy to act as a form of regularization~\cite{chaudhari2019entropy}. Another effective method is to begin training with the Adam optimizer~\cite{kingma2014adam} and later switch to SGD~\cite{robbins1951stochastic}, which has been shown to yield superior generalization performance~\cite{keskar2016large}. Incorporating a partial adaptive parameter to the adaptive gradient methods like Adam and Amsgrad was also introduced~\cite{DBLP:journals/corr/abs-1806-06763}. Label smoothing, which mitigates overconfidence by slightly adjusting target labels~\cite{szegedy2016rethinking}, and FOCA, which discourages co-adaptation between the feature extractor and classifier~\cite{sato2019breaking}, are further examples of techniques designed to enhance generalization.

Another significant area of research investigates how the shape of the loss landscape influences a model’s generalization. Prior studies have demonstrated that both the sharpness of the loss surface and the minimization of the corresponding generalization bounds play a vital role in achieving strong performance across diverse learning tasks~\cite{dziugaite2017computing,keskar2016large,hochreiter1997flat}. Designing optimization algorithms that can effectively discover flatter minima, which could enhance generalization, remains an ongoing challenge. One notable approach addressing this issue is Sharpness-Aware Minimization (SAM)~\cite{foret2020sharpness}, which introduces small perturbations to the model parameters and encourages the model toward flatter regions of the loss landscape. To seek a flat landscape, SAM's optimization process can be categorized into two key steps. First, it identifies a set of parameters (weights) that yields the maximum loss within a small neighborhood around the current parameters. Then, it updates the parameters by minimizing the model loss based on the gradient of loss at this worst-case configuration. SAM has demonstrated strong generalization performance across datasets and model architectures for different tasks~\cite{foret2020sharpness,rat2026flat}. 

Although SAM demonstrates promising performance, its simple objective to minimize the maximum loss within a small neighborhood is not always optimum for obtaining better generalization. Several extensions have been proposed to further improve the performance of SAM. GSAM introduces a new sharpness measurement known as the surrogate gap~\cite{zhuang2022surrogate}. PoF enhances generalization by updating the feature extractor to explore flatter minima~\cite{sato2022pof}. The adaptive sharpness which is scale-invariant was also introduced in ASAM~\cite{DBLP:journals/corr/abs-2102-11600}. GA-SAM is another work that investigates the link between local minima and the generalization ability~\cite{zhang2022ga}. F-SAM decomposes SAM's perturbation into full gradient and stochastic gradient noise components to mitigate the negative effects of the full gradient term~\cite{li2024friendly}. Meanwhile, AACE loss was proposed to replace the standard loss in SAM's perturbation step~\cite{rat2024ada,rat2025imp}.

Despite the remarkable improvement in generalization achieved by SAM and its variants, another major drawback lies in their computational cost. Since SAM requires two backward propagations for each training iteration, it doubles the training time and computational overhead compared to the standard training. This increased computational cost limits its practicality for large-scale models and datasets. For this reason, variants of SAM proposed to apply the SAM algorithm only to some training iterations. For example, SS-SAM uses a Bernoulli trial to randomly apply SAM~\cite{zhao2022ss-sam}. While AE-SAM suggested an adaptive policy to employ SAM based on the geometry of the loss landscape~\cite{jiang2022adaptive}. While SS-SAM and AE-SAM effectively reduce the frequency of applying SAM perturbation steps, they still require some additional backward propagations, which prevents them from fully eliminating the extra computational cost. MSAM, on the other hand, reuses the optimizer’s momentum to define the perturbation direction~\cite{becker2024momentum}. However, MSAM heavily relies on the assumption that the SGD update step overshoots local minima, which may not hold consistently across different training settings, resulting in suboptimal performance.

Another observed limitation in SAM's algorithm is that its perturbation direction solely relies on the raw gradient of the loss of the current mini-batch. Because SAM's perturbation is defined by normalizing the instantaneous gradient, any stochastic variation in the mini-batch directly affects the direction of the perturbation. This makes SAM particularly sensitive to batch-level noise. As a result, the perturbation direction may become unstable or inconsistent across steps, potentially degrading the stability of the training.

To mitigate these limitations while also keeping the robustness of the trained model, in this paper, we propose Exponential Moving Average Sharpness-Aware Minimization (EMASAM). EMASAM replaces SAM’s gradient-based perturbation by integrating Exponential Moving Average (EMA)~\cite{polyak1992accel,ruppert1988efficient} weights, whose parameters are historically smoothed versions of the main model, into the perturbation process. Specifically, for the perturbation step, instead of defining the perturbation direction based on the loss gradient as in the standard SAM, which requires an additional backward pass, EMASAM perturbs the model weights according to the direction from the EMA model, which is a stable reference point, toward the current model, which is less stable. Intuitively, while SAM originally perturbs model weights in the first step seeking the worst-case scenario or the area with the highest loss, EMASAM twists the idea by perturbing from the stable configuration to the area with a less stable setting. Since EMASAM always perturbs the model outward away from the stable position, it can be considered as a softer proxy for the worst-case scenario in the sense that it moves toward a configuration that is worse (less stable) but not necessarily the maximally adverse one. Moreover, EMASAM also avoids gradient-based instability in the perturbation direction because its perturbation direction does not depend on noisy gradients that fluctuate among mini-batches. Instead, EMASAM's perturbation direction always runs away from the globally stable position that is less sensitive to per-batch randomness than raw gradients. Furthermore, after each training step, an EMA shadow copy of the model is created as a smoothed version of the training model by averaging its parameters over time with a decay factor. This EMA counterpart will be used to calculate the perturbation in the next perturbation step. The EMA model is also used for evaluation and inference. 

Consequently, this approach completely eliminates the need for an extra backward pass required in SAM while also preserving the stability of the perturbation direction, keeping the generalization and robustness. Several experiments were conducted to evaluate the effectiveness of the proposed methods. The results across various neural network architectures and benchmark datasets show that EMASAM consistently enhances generalization and overall performance compared to standard SAM while completely eliminating an extra backpropagation.

\section{Preliminary}
\label{sec:pre}
\subsection{Sharpness-Aware Minimization}
\label{ssec: sam}
\vspace*{-2mm}
In conventional deep neural network training, optimization algorithms like Stochastic Gradient Descent (SGD) are employed to minimize the loss function. However, this procedure often converges to sharp minima in the parameter space, regions where the loss is small for training samples but can increase significantly for unseen data. These sharp minima are generally considered less robust and tend to yield inferior generalization compared to flatter solutions.

Sharpness-Aware Minimization (SAM)~\cite{foret2020sharpness} proposes a training paradigm aimed at enhancing the generalization capability of deep neural networks. Conventional optimization methods often converge to sharp minima, leading to weaker generalization. SAM, on the other hands,  searches for parameters that reside in neighborhoods that have uniformly low loss, thus avoiding sharp minima. This objective is formulated as a min–max optimization problem, which can be effectively approximated and optimized using gradient-based methods.

Instead of directly minimizing the loss function as in conventional training, SAM aims to minimize the perturbed loss, which is the loss evaluated at a perturbed configuration, and can be formulated as:
\begin{equation}\label{eq:1}
 L_{\rm SAM}(w) =  \max_{\left\|\varepsilon\right\|\leqslant\rho} L(w+\varepsilon)\,,
\end{equation}
where $L(w)$ is the training loss (typically cross-entropy loss), $w$ represents the model parameters, and $\varepsilon$  is a perturbation vector bounded by $\rho$ in the L2-norm.
The optimization seeks parameters $w$ such that the loss is minimized not just at $w$ but in its neighborhood within a radius of $\rho$.

For the case of small $\rho$, applying Taylor expansion around $w$, the $\varepsilon$ that satisfied the inner maximization in Eq.~\ref{eq:1} can be represented as:

\begin{equation}\label{eq:2}
\varepsilon = {\rm StopGrad} \left( \rho\frac{\triangledown L(w)}{\left\|\triangledown L(w)\right\|_{2}} \right) \,,
\end{equation}
where ${\rm StopGrad}$ represents the stop graduation operation. This formulation identifies the direction in the parameter space where the loss increases most sharply, scaled by the hyperparameter $\rho$. The ${\rm StopGrad}$ operation is included here to ensure that $\varepsilon$ is utilized solely for the perturbation step and remains constant during the computation of gradients for weight updates.

In this paper, we consider Eq.~\ref{eq:2} as a composition of the radius parameter $\rho$ and a specific function responsible for generating the perturbation vector.

\begin{equation}\label{eq:3}
\varepsilon = {\rm StopGrad}(\rho \, g(w))\,.
\end{equation}
where $g(w)$ is named a perturbation-generating function. In SAM, this function is described as
\begin{equation}\label{eq:4}
g_{\rm SAM}(w) = \frac{\triangledown L(w)}{\left\|\triangledown L(w)\right\|_{2}}\,,
\end{equation}

SAM's algorithm consists of two main steps. First, the algorithm perturbs the model parameters toward the worst-case direction using the perturbation vector defined in Eq. \ref{eq:2}. For the second step, at the original position, SAM updates the model weights by optimizing the model parameters using the gradients of the loss calculated at the perturbed position, as expressed in the following equation.

\begin{equation}\label{eq:5}
w_{\rm t+1} = w_{\rm t}-\eta\triangledown L(w_{\rm t}+ \varepsilon)\,,
\end{equation}
where $\eta$ is the learning rate. Note that for simplicity's sake, this weight updating formula is based on standard SGD without the momentum. Nonetheless, in practical implementations, other optimization methods such as Adam, RMSprop, or SGD with momentum can also be applied.

This optimization strategy promotes convergence toward flatter minima, which are generally associated with improved generalization to unseen data. SAM has been shown to enhances the performance of a wide range of deep learning models across tasks such as image classification and face recognition.

\subsection{Exponential Moving Average Model}
\label{ssec: ema}
\vspace*{-2mm}
The Exponential Moving Average (EMA)~\cite{polyak1992accel,ruppert1988efficient} has become a widely adopted technique in deep learning to enhance training stability and improve generalization. EMA model helps filter out noise in parameter updates that often arise during stochastic optimization.

The core idea of the EMA model is to maintain a shadow model whose parameters are an exponentially weighted average of the parameters of the current training model. This shadow model is updated at the end of each training iteration, following the optimizer’s update step. By averaging parameters over multiple iterations, the EMA model captures a more stable representation that typically yields better performance during inference. The intuition is that while the instantaneous model parameters may overfit to the most recent mini-batch or display high variance, the EMA parameters represent a temporally smoothed version that approximates the ensemble effect of multiple recent models without increasing computational cost. The EMA model updating can be represented as

\begin{equation}\label{eq:6}
w_{\rm t+1}^{\rm EMA} = \alpha w_{\rm t}^{\rm EMA} + (1-\alpha)w_{\rm t+1}\,,
\end{equation}
where $w_{\rm t}^{\rm EMA}$ denotes the EMA parameters at iteration $t$, $w_{\rm t+1}$ denotes the parameters of the main model after the weight updating step, and $\alpha \in [0,1]$ is the decay rate that controls how strongly past parameters influence the average. In practice, during evaluation or inference, this EMA model is often used instead of the raw model parameters, as it tends to achieve better test accuracy and robustness to noise.

Overall, the EMA model serves as a cheap yet effective mechanism to approximate a more stable point in the loss landscape, allowing the model to obtain a smoother and more generalizable solution without modifying the optimization objective or requiring additional backward passes.

\section{Proposed Method}
\label{sec:propose}

\begin{figure*}[!t]
    \centering
    % First row
    \subfloat[SGD.]{%
        \includegraphics[width=0.45\textwidth]{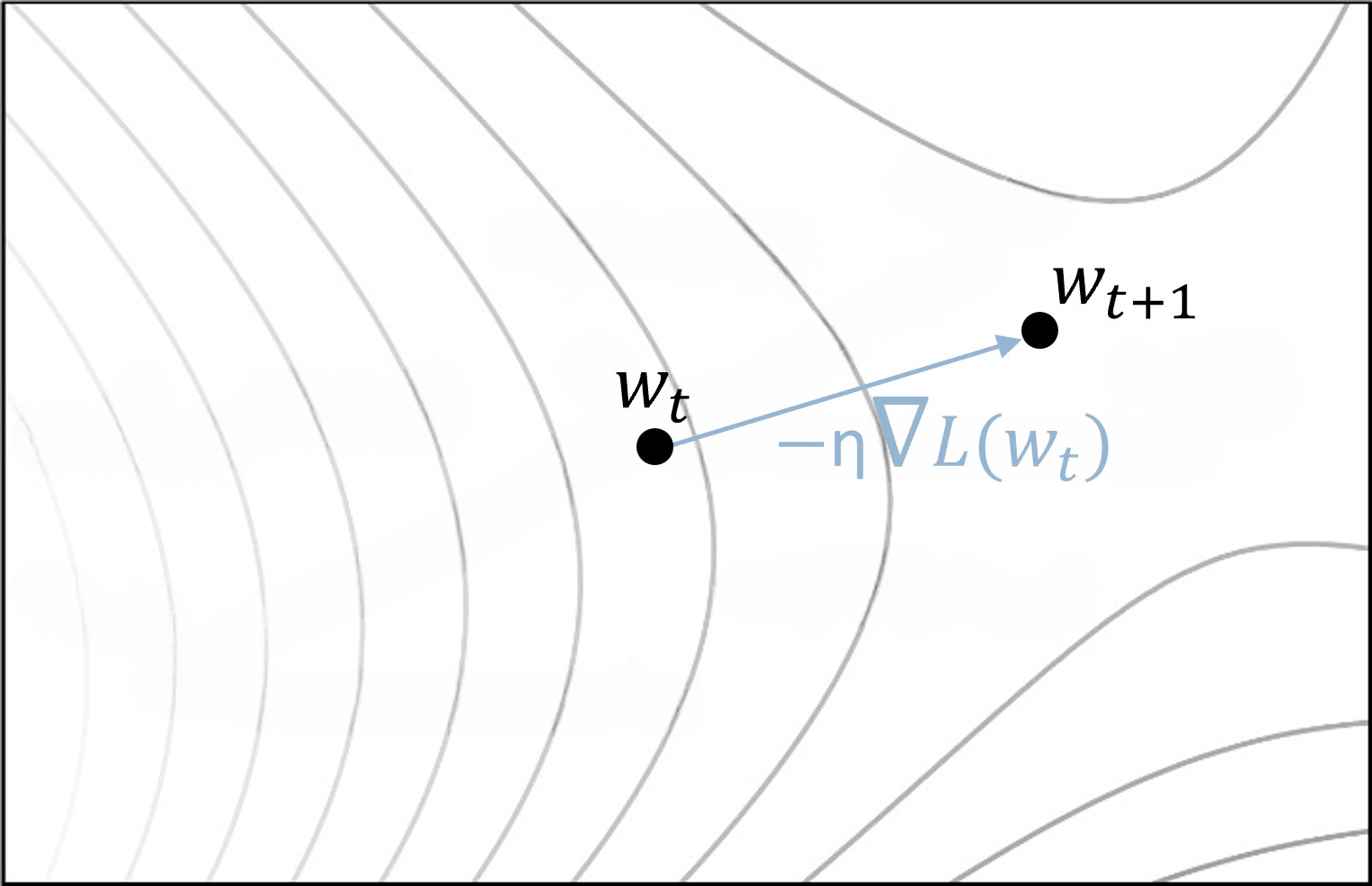}%
    }%
    \hfill
    \subfloat[EMA.]{%
        \includegraphics[width=0.45\textwidth]{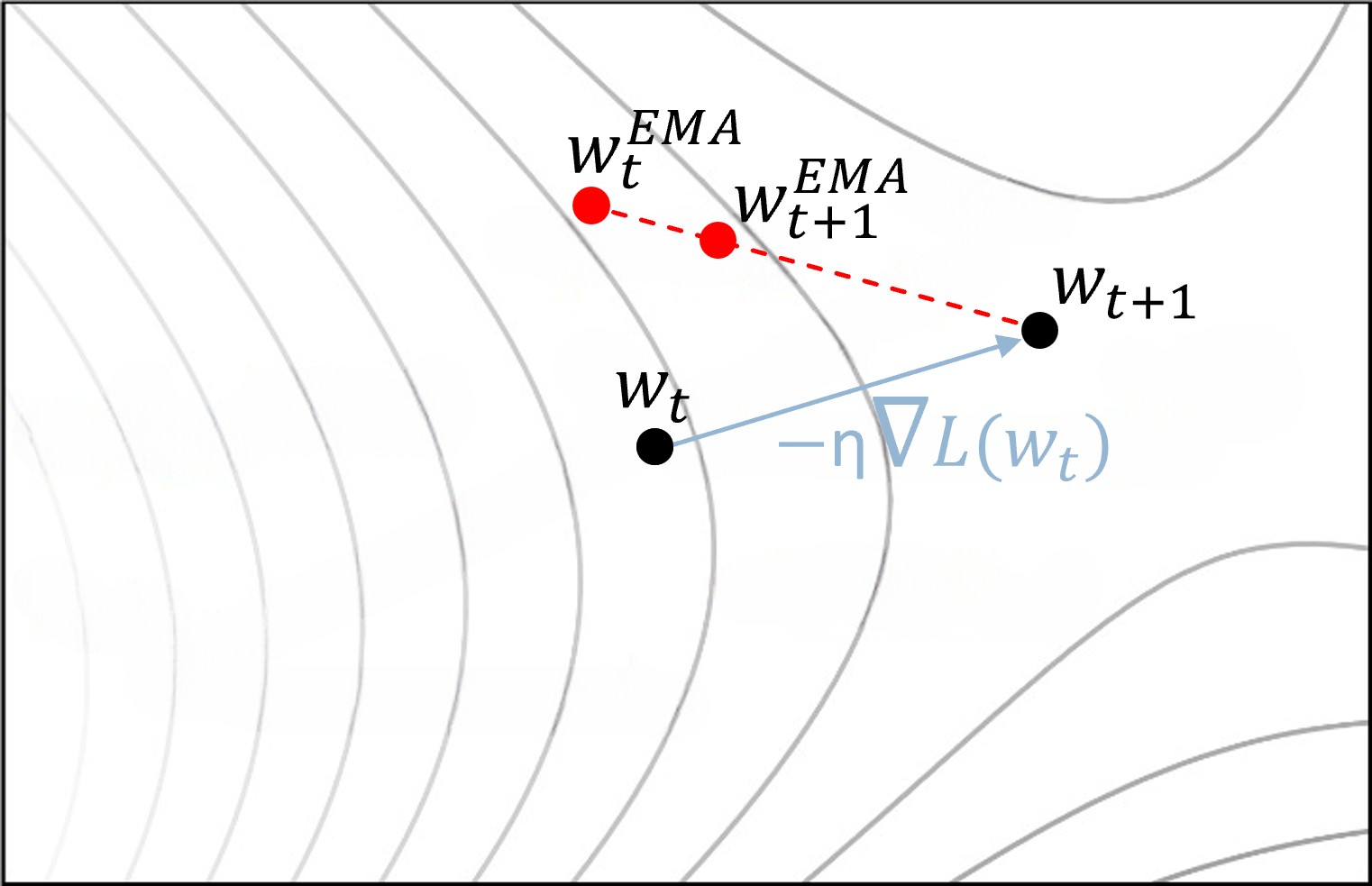}%
    }\\[1ex] % line break between rows

    % Second row
    \subfloat[SAM.]{%
        \includegraphics[width=0.45\textwidth]{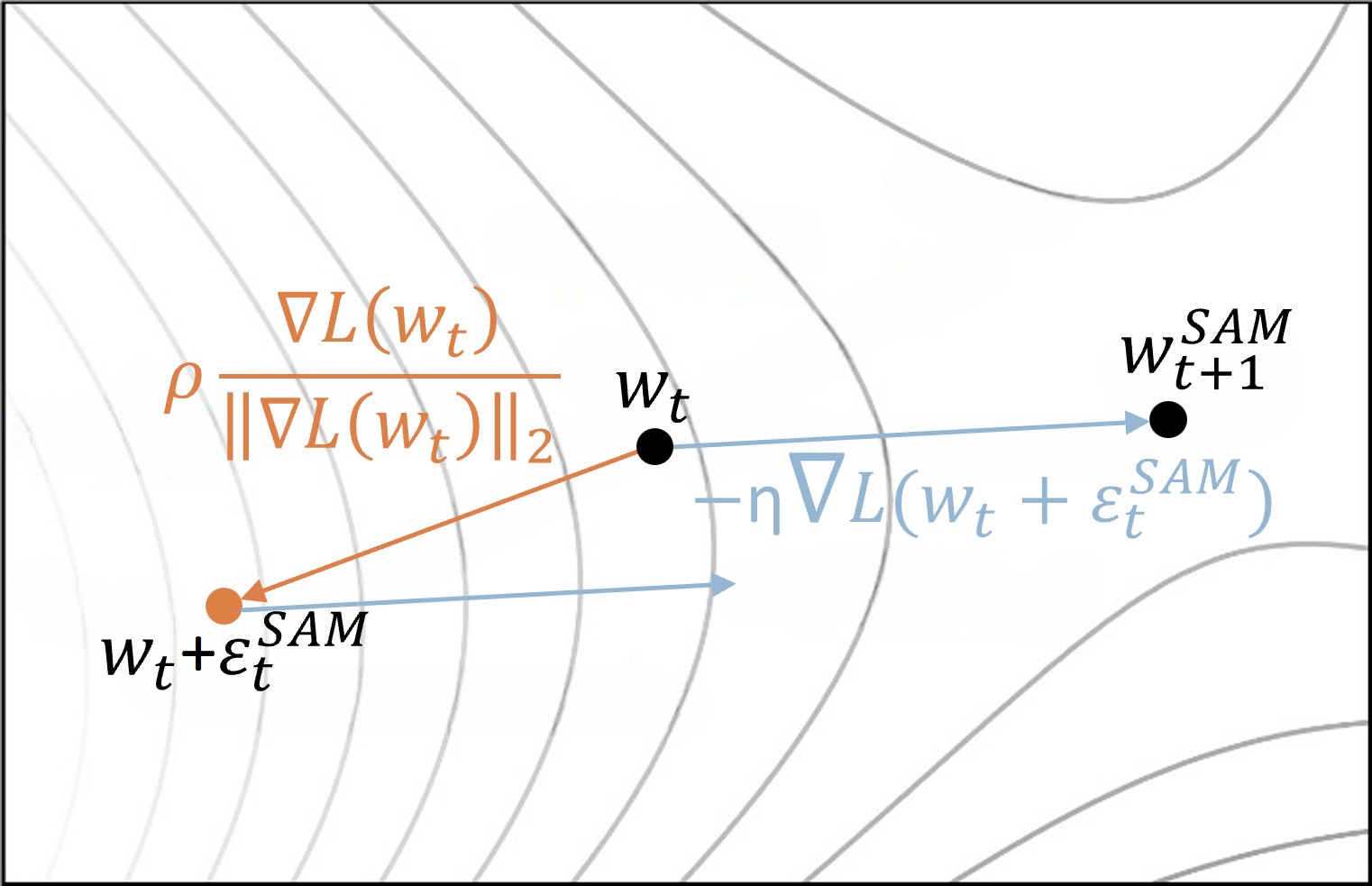}%
    }%
    \hfill
    \subfloat[EMASAM]{%
        \includegraphics[width=0.45\textwidth]{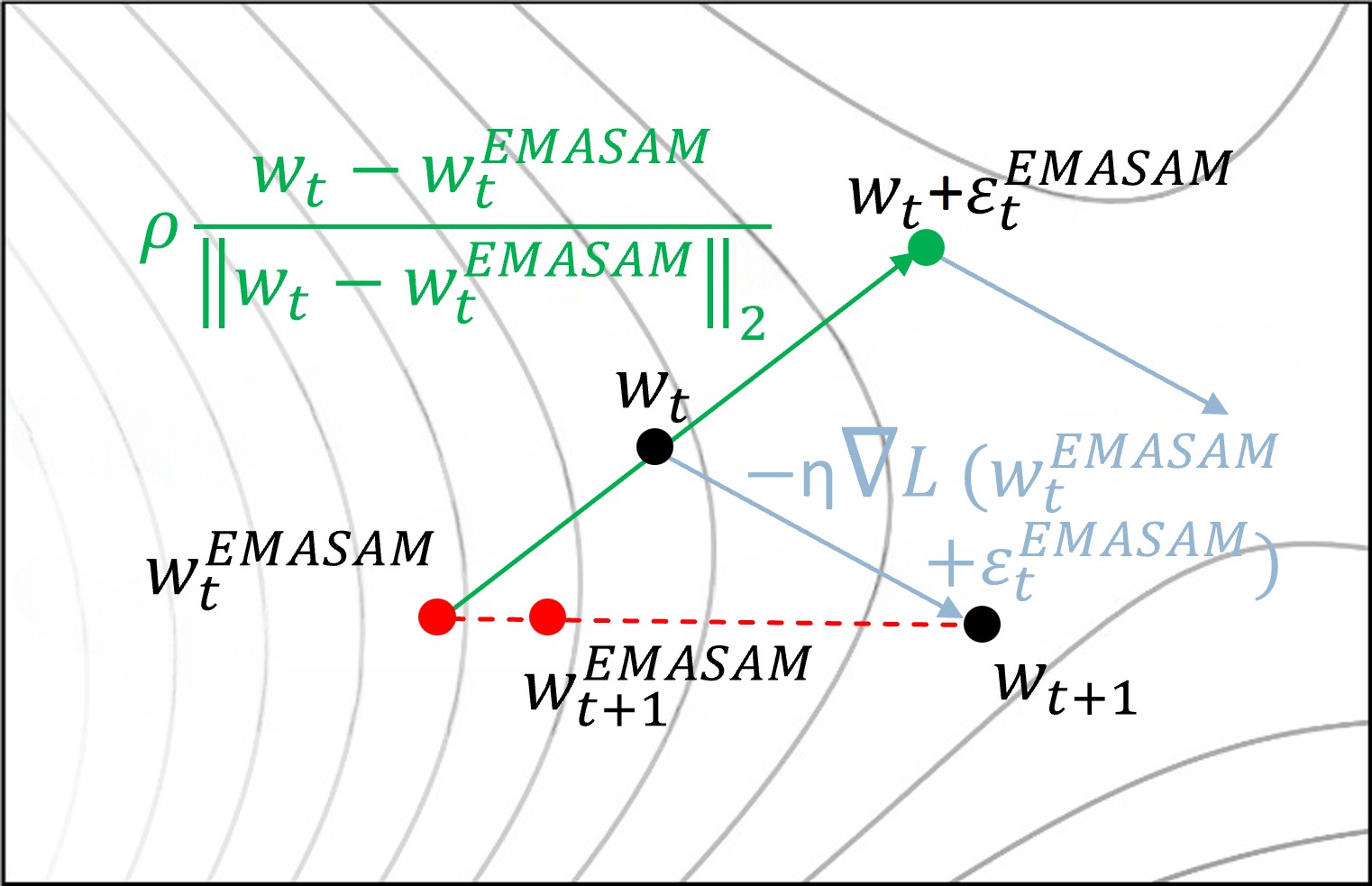}%
    }
    \vspace*{-2mm}
    \caption{Diagram illustrates the algorithm of SGD, EMA, SAM, and EMASAM.}
    \label{fig:1}
    \vspace*{-5mm}
\end{figure*}

Although SAM exhibits strong generalization performance, certain aspects of its perturbation mechanism remain underexplored. Specifically, SAM determines the worst-case direction by constructing a perturbation vector based on the normalized gradient of the training loss, scaled by a fixed radius that defines the local neighborhood, as shown in Eq.~\ref{eq:2}.
This leads to two main limitations. First, it requires an additional backward propagation. As a result, SAM requires two backward propagations per training iteration, making it suffer from a considerable computational burden that doubles both the training time and computational cost compared to standard optimization methods. This overhead limits its scalability to large models and datasets. Another limitation of SAM is that its perturbation direction depends entirely on the raw gradient computed from the current mini-batch. This makes SAM highly sensitive to batch-level noise, causing the perturbation direction to vary significantly across iterations. Such variability can lead to instability in the perturbation step and may negatively impact the overall training dynamics.

To address these limitations while preserving the robustness of SAM, we propose Exponential Moving Average Sharpness-Aware Minimization (EMASAM), a computationally efficient variant of SAM. EMASAM replaces SAM’s gradient-based perturbation mechanism with one that leverages the Exponential Moving Average (EMA) model, whose parameters serve as a temporally smoothed and more stable counterpart to the main model.

In the perturbation step, EMASAM does not rely on the loss gradient, which in SAM requires an additional gradient computation. Instead, EMASAM perturbs the parameters along the direction from the EMA model, which is a stable reference point, toward the current model, which is inherently less stable. Mathematically, during the perturbation step, EMASAM define the perturbation-generating function as
\begin{equation}\label{eq:7}
g_{\rm EMASAM}(w) = \frac{(w-w^{\rm EMASAM})}{\left\|w-w^{\rm EMASAM}\right\|_{2}}\,,
\end{equation}
where $(w-w^{\rm EMASAM})$ represents a vector pointing from the EMA shadow model toward the trained model. Intuitively, this vector perturbs the model from the more stable position toward the less stable position. In other words, it serves as a softer approximation of the worst-case direction as the perturbation moves toward a less favorable configuration rather than the maximally adversarial one. Now that the perturbation direction is defined by the discrepancy between the main model and the EMA model, the extra backpropagation is not required.

In addition, EMASAM mitigates the gradient-induced instability inherent in SAM. Because EMASAM’s perturbation direction does not depend on noisy mini-batch gradients, it avoids the fluctuations that arise from batch-to-batch randomness. Instead, the direction is consistently defined relative to a globally referenced baseline, a smoothed EMA position, making it substantially more stable than perturbations derived from instantaneous raw gradients.

For the second step, the gradient for the weight updating is computed at the perturbed position. Then, from the original parameter position, the model weights are updated by optimizing with respect to this calculated gradient. 

The main model updating rule can now be presented as
\begin{equation}\label{eq:8}
w_{t+1} = w_{t}-\eta\triangledown L(w^{\rm EMASAM}_t+\varepsilon^{\rm EMASAM}_t)\,,
\end{equation}
where
\begin{equation}\label{eq:9}
\varepsilon^{\rm EMASAM}_t = {\rm StopGrad} \left(\rho \frac{(w_t-w^{\rm EMASAM}_t)}{\left\|w_t-w^{\rm EMASAM}_t\right\|_{2}}\ \right)\,.
\end{equation}

Moreover, after each update, an EMA shadow model is maintained as a temporally smoothed version of the training model, where parameters are averaged over time using a decay coefficient. Hence, this can be expressed as
\begin{equation}\label{eq:10}
w_{\rm t+1}^{\rm EMASAM} = \alpha w_{\rm t}^{\rm EMASAM} + (1-\alpha)(w_{t}-\eta\triangledown L(w^{\rm EMASAM}_t+\varepsilon^{\rm EMASAM}_t))\,.
\end{equation}

The comparison of these weight-updating algorithms is shown in Fig.~\ref{fig:1}. SGD updates the model parameters directly using the gradient of the loss at the current position. While the standard EMA model follows the optimizer's step by updating a shadow model whose parameters are a weighted average of the parameters of the current training model. SAM add a small perturbation to the weights, then the gradient is computed at this perturbed position, and this gradient is used to update the weights at the original configuration. Our approach, EMASAM, also has a small perturbation step. However, while SAM requires additional backpropagation in order to obtain the gradient for the perturbation defining, EMASAM defines the perturbation vector based on the difference between the training model and the EMA counterpart, requiring no extra backpropagation. At the end, our method also maintain the shadow copy after the weight updating step, similar to the EMA model.

As a result, the proposed method no longer requires additional gradient computation and also benefits from a more stable perturbation direction. This helps maintain generalization and robustness without extra computational overhead.

\begin{algorithm}[t]
\caption{EMASAM}
\SetAlgoLined
\DontPrintSemicolon

\KwIn{training data $\mathcal{S}$, learning rate $\eta$, perturbation radius $\rho$, EMA coefficient $\alpha$}
Initialize weights $w_0$, EMA weights $w_0^{\rm EMASAM}$

\For{$t = 0$ \KwTo $T-1$}{
    Sample mini-batch $\mathcal{B}_t \subset \mathcal{S}$\;

    Perturb weight: $w^{\rm EMASAM}_t+\varepsilon^{\rm EMASAM}_t \leftarrow w^{\rm EMASAM}_t + \rho \frac{(w_t-w_t^{\rm EMASAM})}{\left\|w_t-w_t^{\rm EMASAM}\right\|_{2}}\ $

    Compute gradient: $\triangledown L(w^{\rm EMASAM}_t+\varepsilon^{\rm EMASAM}_t)$\;
    Update main model: $w_{t+1} \leftarrow w_{t}-\eta\triangledown L(w^{\rm EMASAM}_t+\varepsilon^{\rm EMASAM}_t)$\;
    Update EMASAM model: $w_{\rm t+1}^{\rm EMASAM} \leftarrow \alpha w_{\rm t}^{\rm EMASAM} + (1-\alpha)w_{t+1}$
}
\Return{$w_{t+1}, w_{\rm t+1}^{\rm EMASAM}$}\;
\end{algorithm}

\section{Experiments}
\label{sec:experiments}
% To evaluate the practical effectiveness of the proposed method, we conducted a series of empirical experiments.

\subsection{Hyperparameter Tuning}
\label{ssec: grid search}
\vspace*{-2mm}

\begin{table}[t!]
\caption{Accuracies ($\%$) for hyperparameter tuning of EMASAM on ResNet-50 on CIFAR-100}
\centering
\begin{tabular}{ |c|c|c| } 
 \hline
   $\rho$ & with normalization & w/o normalization\\ 
 \hline
0.01  & 81.74$\pm$0.24 & 80.85$\pm$0.33\\ 
0.02  & 81.44$\pm$0.13 & 81.03$\pm$0.39\\ 
0.05  & 81.46$\pm$0.26 & 80.55$\pm$0.17\\ 
0.1   & 81.26$\pm$0.47 & 81.38$\pm$0.24\\ 
0.2   & 81.24$\pm$0.17 & 81.15$\pm$0.15\\ 
0.5   & 81.65$\pm$0.32 & 80.58$\pm$0.58\\ 
1.0   & 81.74$\pm$0.37 & 81.46$\pm$0.18\\ 
2.0   & \textbf{82.34$\pm$0.54} & 81.71$\pm$0.37\\ 
5.0   & \underline{81.82}$\pm$0.17 & 81.54$\pm$0.24\\ 
 \hline
\end{tabular}
\vspace*{-6mm}
\label{table:1}
\end{table}

To assess the performance of our EMASAM, we conducted initial experiments using the ResNet-50~\cite{he2016deep} architecture on the CIFAR-100~\cite{krizhevsky2009learning} dataset. SGD with a weight decay of 0.0005 was employed as the base optimizer. It is worth noting that momentum was not used in the updating for EMASAM, as incorporating momentum may excessively smooth the EMASAM dynamics. Horizontal flip, padding by four pixels, random crop and cutout regularization~\cite{devries2017improved} were applied for data augmentations. Label smoothing was also applied with a smoothing factor = 0.1. The momentum for running statistics in batch normalization was disabled as they would be updated during the EMA process. The decay rate for the EMA model update, $\alpha$, was set to 0.99. The models were trained from scratch with a batch size = 32 for 200 epochs. We set the initial learning rate to 0.1 and drop by 0.2 at 30\%, 60\%, and 80\% of the training. The perturbation scaler $\rho$, was tuned over \{0.01, 0.02, 0.05, 0.1, 0.2, 0.5, 1.0, 2.0, 5.0\}. We also tried the ablation study on the perturbations without normalization, which has a similar strategy to the Schedule-Free SGD~\cite{defazio2024road}. Each setting was repeated three times. 
The average results with standard deviation in Table 1 shows that the perturbation with normalization consistently outperform those without normalization, confirming the importance of normalization in the perturbation-generating function.

More importantly, we realized that additional sensitivity to hyperparameter choices, $\rho$, is important to obtain better accuracy in practice.
%can increase the complexity of deploying a method in practice. 
During our hyperparameter tuning as showing Table~\ref{table:1}, the setting where $\rho = 2.0$ yielded the strongest performance. Hence, to demonstrate that our approach does not require re-tuning of this parameter for every single configurations, we fixed $\rho = 2.0$ for all subsequent experiments involving EMASAM.

\subsection{Image Classification Performances on CIFAR-10, CIFAR-100, Fashion-MNIST, and EMNIST Datasets}
\vspace*{-2mm}
To validate the effectiveness of our EMASAM, we performed empirical evaluations using the various network architectures across multiple datasets, such as CIFAR-10, CIFAR-100~\cite{krizhevsky2009learning}, Fashion-MNIST~\cite{xiao2017fashion}, and EMNIST~\cite{cohen2017emnist}. We also conducted the experiments using other methods. The standard SGD~\cite{robbins1951stochastic} was used as a baseline. SAM~\cite{foret2020sharpness} and its variants, SS-SAM~\cite{zhao2022ss-sam}, AE-SAM~\cite{jiang2022adaptive}, and MSAM~\cite{becker2024momentum} were included in the experiments. The conventional EMA model was also included to compare. Lastly, to confirm the importance of the perturbation direction, we tried replacing EMASAM's perturbation direction with a random vector, creating the EMA model with a random perturbation. Each setting was repeated three times. The average results with standard deviations are reported.

\begin{table*}[t!]
\caption{Accuracies ($\%$) of models trained with ResNet-18}
\centering
\begin{tabular}{ |c|c|c|c|c|c| } 
\hline
Method & CIFAR-10 & CIFAR-100 & Fashion-MNIST & EMNIST & Average\\
\hline
Vanilla (SGD) & 94.57$\pm$0.07 & 77.62$\pm$0.47 & 89.81$\pm$0.78 & 88.07$\pm$0.24 & 87.52\\ 
SAM~\cite{foret2020sharpness} & \underline{96.09$\pm$0.03} & \underline{79.80$\pm$0.29} & 92.21$\pm$0.14 & \textbf{90.11$\pm$0.02} & \underline{89.55}\\ 
SS-SAM~\cite{zhao2022ss-sam} & 95.64$\pm$0.31 & 79.34$\pm$0.85 & 91.93$\pm$0.74 & 89.75$\pm$0.17 & 89.17\\ 
AE-SAM & 95.79$\pm$0.08 & 78.71$\pm$0.55 & \underline{92.39$\pm$0.19} & 90.02$\pm$0.12 & 89.23\\ 
MSAM~\cite{becker2024momentum} & 94.84$\pm$0.14 & 78.18$\pm$0.16 & 92.21$\pm$0.21 & 89.39$\pm$0.48 & 88.65\\ 
EMA & 96.04$\pm$0.11 & 79.25$\pm$0.25 & 90.98$\pm$0.14 & 88.01$\pm$0.15 & 88.57\\ 
EMA + RandPerturb & 93.34$\pm$0.03 & 74.79$\pm$0.13 & 91.60$\pm$0.16 & 89.93$\pm$0.11 & 87.42\\ 
EMASAM [Ours] & \textbf{96.23$\pm$0.09} & \textbf{79.87$\pm$0.17} & \textbf{93.36$\pm$0.11} & \underline{90.08$\pm$0.10} & \textbf{89.88}\\ 
\hline
\end{tabular}
\vspace*{-6mm}
\label{table:2}
\end{table*}

\begin{table*}[t!]
\caption{Accuracies ($\%$) of models trained with ResNet-50}
\centering
\begin{tabular}{ |c|c|c|c|c|c| } 
\hline
Method & CIFAR-10 & CIFAR-100 & Fashion-MNIST & EMNIST & Average\\ 
\hline
Vanilla (SGD) & 94.56$\pm$0.15 & 77.21$\pm$0.09 & 90.87$\pm$1.03 & 88.44$\pm$0.16 & 87.77\\ 
SAM~\cite{foret2020sharpness} & 96.00$\pm$0.13 & 79.41$\pm$0.61 & \underline{92.33$\pm$0.24} & 89.25$\pm$0.60 & 89.25\\ 
SS-SAM~\cite{zhao2022ss-sam} & 95.63$\pm$0.18 & 79.91$\pm$0.26 & 92.22$\pm$0.00 & 89.43$\pm$0.11 & \underline{89.30}\\ 
AE-SAM~\cite{jiang2022adaptive} & 95.45$\pm$0.33 & 78.82$\pm$1.32 & 92.18$\pm$0.09 & 89.02$\pm$0.11 & 88.87\\ 
MSAM~\cite{becker2024momentum} & 93.16$\pm$0.21 & 78.16$\pm$0.73 & 90.41$\pm$0.24 & 88.41$\pm$0.53 & 87.54\\ 
EMA & \underline{96.49$\pm$0.06} & \underline{81.50$\pm$0.23} & 90.38$\pm$0.10 & 87.96$\pm$0.09 & 89.08\\ 
EMA + RandPerturb & 92.92$\pm$0.03 & 74.90$\pm$0.46 & 91.20$\pm$0.07 & \underline{89.61$\pm$0.07} & 87.16\\ 
EMASAM [Ours] & \textbf{96.78$\pm$0.20} & \textbf{82.34$\pm$0.54} & \textbf{92.57$\pm$0.10} & \textbf{89.91$\pm$0.01} & \textbf{90.40}\\ 
\hline
\end{tabular}
\vspace*{-6mm}
\label{table:3}
\end{table*}

\begin{table*}[t!]
\caption{Accuracies ($\%$) of models trained with WideResNet-28-10}
\centering
\begin{tabular}{ |c|c|c|c|c|c| } 
\hline
Method & CIFAR-10 & CIFAR-100 & Fashion-MNIST & EMNIST & Average\\ 
\hline
Vanilla (SGD) & 95.83$\pm$0.11 & 81.50$\pm$0.27 & 89.54$\pm$0.88 & 86.67$\pm$0.38 & 88.39\\ 
SAM~\cite{foret2020sharpness} & \underline{96.96$\pm$0.05} & \textbf{84.05$\pm$0.27} & 91.63$\pm$0.26 & \underline{89.58$\pm$0.13} & \underline{90.55}\\ 
SS-SAM~\cite{zhao2022ss-sam} & 96.75$\pm$0.15 & \underline{83.03$\pm$0.09} & 91.54$\pm$0.26 & \textbf{89.74$\pm$0.15} & 90.26\\ 
AE-SAM~\cite{jiang2022adaptive} & \underline{96.96$\pm$0.08} & 83.40$\pm$0.10 & \underline{91.89$\pm$0.40} & 89.55$\pm$0.13 & 90.45\\ 
MSAM~\cite{becker2024momentum} & 95.74$\pm$0.02 & 82.27$\pm$0.13 & 91.63$\pm$0.40 & 87.89$\pm$0.32 & 89.38\\ 
EMA & 96.70$\pm$0.01 & 81.96$\pm$0.02 & 90.39$\pm$0.23 & 86.10$\pm$0.19 & 88.79\\ 
EMA + RandPerturb & 94.73$\pm$0.06 & 77.75$\pm$0.17 & 91.76$\pm$0.03 & 90.03$\pm$0.11 & 88.57\\ 
EMASAM [Ours] & \textbf{97.13$\pm$0.13} & 82.96$\pm$0.04 & \textbf{92.94$\pm$0.09} & 89.42$\pm$0.02 & \textbf{90.61}\\ 
\hline
\end{tabular}
\vspace*{-6mm}
\label{table:4}
\end{table*}

\begin{table*}[t!]
\caption{Accuracies ($\%$) of models trained with PyramidNet-110}
\centering
\begin{tabular}{ |c|c|c|c|c|c| } 
\hline
Method & CIFAR-10 & CIFAR-100 & Fashion-MNIST & EMNIST & Average\\ 
\hline
Vanilla (SGD) & 94.08$\pm$0.04 & 75.92$\pm$0.39 & 87.84$\pm$1.00 & 89.03$\pm$0.20 & 86.72\\ 
SAM~\cite{foret2020sharpness} & 95.91$\pm$0.04 & 79.15$\pm$0.52 & \underline{91.53$\pm$0.30} & 89.16$\pm$0.27 & 88.92\\ 
SS-SAM~\cite{zhao2022ss-sam} & 95.49$\pm$0.25 & 80.19$\pm$0.16 & 91.02$\pm$0.96 & 88.41$\pm$0.28 & 88.58\\ 
AE-SAM~\cite{jiang2022adaptive} & 96.10$\pm$0.46 & 79.22$\pm$0.90 & 91.47$\pm$0.33 & 89.38$\pm$0.10 & \underline{89.04}\\  
MSAM~\cite{becker2024momentum} & 93.61$\pm$0.35 & 77.49$\pm$0.25 & 90.32$\pm$0.30 & 88.39$\pm$0.26 & 87.45\\ 
EMA & \underline{96.58$\pm$0.05} & \underline{80.74$\pm$0.41} & 89.48$\pm$0.38 & 86.78$\pm$0.21 & 88.40\\ 
EMA + RandPerturb & 93.63$\pm$0.11 & 73.19$\pm$0.19 & 91.02$\pm$0.01 & \textbf{89.86$\pm$0.07} & 86.22\\ 
EMASAM [Ours] & \textbf{96.77$\pm$0.05} & \textbf{81.86$\pm$0.27} & \textbf{92.56$\pm$0.10} & \underline{89.69$\pm$0.06} & \textbf{90.22}\\ 
\hline
\end{tabular}
\vspace*{-6mm}
\label{table:5}
\end{table*}

\begin{table}[t!]
\caption{Accuracies ($\%$) of models trained with ResNet-50 on ImageNet-1k}
\centering
\begin{tabular}{ |c|c|c| } 
 \hline
   Method & Top-1 & Top-5\\ 
 \hline
Vanilla (SGD) & 76.45 & 93.22 \\ 
SAM~\cite{foret2020sharpness} & \underline{76.71} & 93.32 \\ 
SS-SAM~\cite{zhao2022ss-sam} & 76.64 & \textbf{93.48} \\ 
AE-SAM~\cite{jiang2022adaptive} & 76.68 & \underline{93.41} \\ 
MSAM~\cite{becker2024momentum} & 75.83 & 92.87 \\ 
EMA & 76.51 & 92.87 \\  
EMA + RandPerturb & 60.58 & 83.07 \\ 
EMASAM [Ours]& \textbf{76.87} & 93.23 \\
 \hline
\end{tabular}
\vspace*{-6mm}
\label{table:6}
\end{table}

\begin{table}[t!]
\caption{Accuracies ($\%$) of models trained with ViT-B/16 on ImageNet-1k}
\centering
\begin{tabular}{ |c|c|c| } 
 \hline
   $Method$ & Top-1 & Top-5\\ 
 \hline
Vanilla (AdamW) & 69.16 & 87.79 \\ 
SAM~\cite{foret2020sharpness} & 71.19 & 89.21 \\ 
SS-SAM~\cite{zhao2022ss-sam} & 69.58 & 88.28 \\  
AE-SAM~\cite{jiang2022adaptive} & \textbf{72.06} & \underline{89.97} \\ 
MSAM~\cite{becker2024momentum} & 71.59 & 89.48 \\
EMA & 71.04 & 89.42 \\ 
EMA + RandPerturb & 46.67 & 70.12 \\ 
EMASAM [Ours]& \underline{72.00} & \textbf{90.15} \\ 
 \hline
\end{tabular}
\vspace*{-6mm}
\label{table:7}
\end{table}

As for the experiment setting, an SGD optimizer with a weight decay of 0.0005 was used as the base optimizer. The optimizer momentum was set to 0 for EMASAM and its random perturbation variant, and 0.9 for all other methods. Horizontal flip, padding by four pixels, random crop, cutout regularization, and Label smoothing with the smoothing factor = 0.1 were also applied. The EMA decay was set to 0.99. The models were trained from scratch with a batch size = 32 for 200 epochs. The initial learning rate was set to 0.1 and drop by 0.2 at 30\%, 60\%, and 80\% of the training. We kept $\rho = 2.0$ for EMASAM, and used $\rho = 3.0$ for MSAM, following the optimal value reported in the MSAM paper. For the standard SAM, SS-SAM, and AE-SAM that define perturbation in the same way, we used $\rho = 0.05$ as suggested in SAM's original paper. In the SS-SAM implementation, the probability of performing SAM, $a_c$, was fixed at 0.5. While the $\lambda_1$ and $\lambda_2$ for AE-SAM were set to -1 and 1, respectively.

% \subsubsection{Image Classification Comparisons with ResNet-18}
First, we conducted the experiments on ResNet-18~\cite{he2016deep}. We report the average accuracies with the standard deviation in Table 2. As shown in Table 2, our method achieves the highest test accuracies on most datasets, including CIFAR-10, CIFAR-100, and Fashion-MNIST. For EMNIST, our method ranks second, slightly behind standard SAM. EMASAM also has the highest average accuracy among all datasets used in our experiments.

% \subsubsection{Image Classification Comparisons with ResNet-50}
We also conducted the experiments on ResNet-50~\cite{he2016deep}. The empirical results are reported in Table 3. As seen in Table 3, our methods perform best on all datasets. EMASAM also achieves the best average accuracy among all datasets.

% \subsubsection{Image Classification Comparisons with WideResNet-28-10}
The experiments were also conducted on the Wide ResNet~\cite{zagoruyko2016wide} model. We used model depth = 28, width factor = 10 for these experiments. As presented in Table 4, our method performs best on CIFAR-10 and Fashion-MNIST, although it lags behind SAM and SS-SAM on CIFAR-100 and EMNIST. However, EMASAM still ranks first in the average accuracy.

% \subsubsection{Image Classification Comparisons with PyramidNet-110}
We also carried out experiments using a PyramidNet-110~\cite{han2017deep} model, configured with a depth = 110 and an alpha = 84. The results are shown in Table 5. EMASAM achieves the highest accuracies on CIFAR-10, CIFAR-100, and Fashion-MNIST, while attaining the second-highest accuracy on the EMNIST dataset. EMASAM also has the highest average accuracy.

\subsection{Image Classification Performances on ImageNet Datasets}
\vspace*{-2mm}
To further demonstrate the effectiveness of our method on large-scale applications, we conducted experiments on the ImageNet-1K~\cite{deng2009imagenet} dataset using different networks, including Vision Transformer~\cite{dosovitskiy2020image} and ResNet~\cite{he2016deep}. The training set was prepared using Inception-style preprocessing~\cite{szegedy2015going}. Following our previous experiments, we set $\rho$ = 2.0 for EMASAM, 3.0 for MSAM, and 0.05 for SAM and variants with the same perturbation method.

% \subsubsection{Image Classification Comparisons with ResNet-50}
Firstly, ResNet-50 was selected as a model architecture. We used SGD with momentum = 0.9 (0.0 for EMASAM and EMA with random perturbation), and weight decay = 0.0001 as a base optimizer. The models were trained from scratch with a batch size = 1,024 for 100 epochs. We used cosine decay learning rate with 20,000 warmup steps, and a peak learning rate was set to 1.0. The experiment results in Table 6 confirm the best robustness of EMASAM for top-1 accuracy although it is not the best for top-5 accuracy.

% \subsubsection{Image Classification Comparisons with ViT-B/16}
For the Vision Transformer, we used ViT-B/16 as our model. AdamW was used as a base optimizer with the exponential decay rates set to (0.0, 0.999) for EMASAM and EMA with random perturbation, and set to (0.9, 0.999) for other methods. The weight decay was set to 0.3. We trained the models from scratch for 300 epochs with a batch size = 512. We used a cosine decay learning rate with 80,000 warmup steps, and a peak learning rate was set to 3e-3. We report the experiment results in Table 7. As shown in Table 7, our method achieves the best top-5 accuracy while ranking best at top-1 accuracy, confirming the robustness of the proposed EMASAM.\\

The experiments in sections 4.2 and 4.3 show that EMASAM often matches or outperforms the conventional EMA model, the standard SAM model, and other SAM variants. Moreover, it can be noticed from the results that the EMA with a random perturbation direction significantly degraded the robustness of the model compared to EMASAM. This confirms the importance of perturbation direction and the effectiveness of the newly defined perturbation in EMASAM.

% \begin{figure}
%     \centering
%     \includegraphics[height=5.8cm]{BackProp_ResNet50_CIFAR100.jpg}
%     \caption{Number of backpropagation and test accuracy comparison among models trained with ResNet-50 on CIFAR-100.}
%     \label{fig:2}
% \end{figure}

% \begin{figure}
%     \centering
%     \includegraphics[height=5.8cm]{BackProp_ResNet50_ImageNet.jpg}
%     \caption{Number of backpropagation and test accuracy comparison among models trained with ResNet-50 on ImageNet-1k.}
%     \label{fig:3}
% \end{figure}

\begin{figure*}[!t]
    \centering
    % First row
    \subfloat[ResNet-50 on CIFAR-100]{%
        \includegraphics[width=0.4\textwidth]{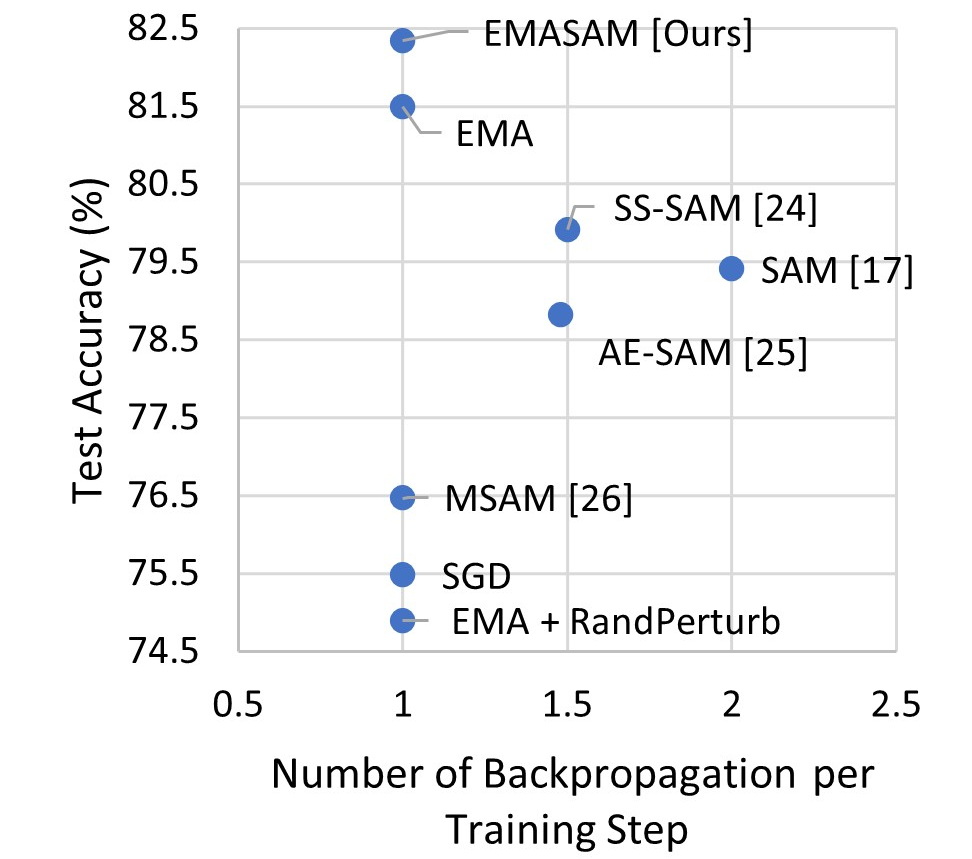}%
    }%
    \hfill
    \subfloat[ResNet-50 on ImageNet-1k]{%
        \includegraphics[width=0.4\textwidth]{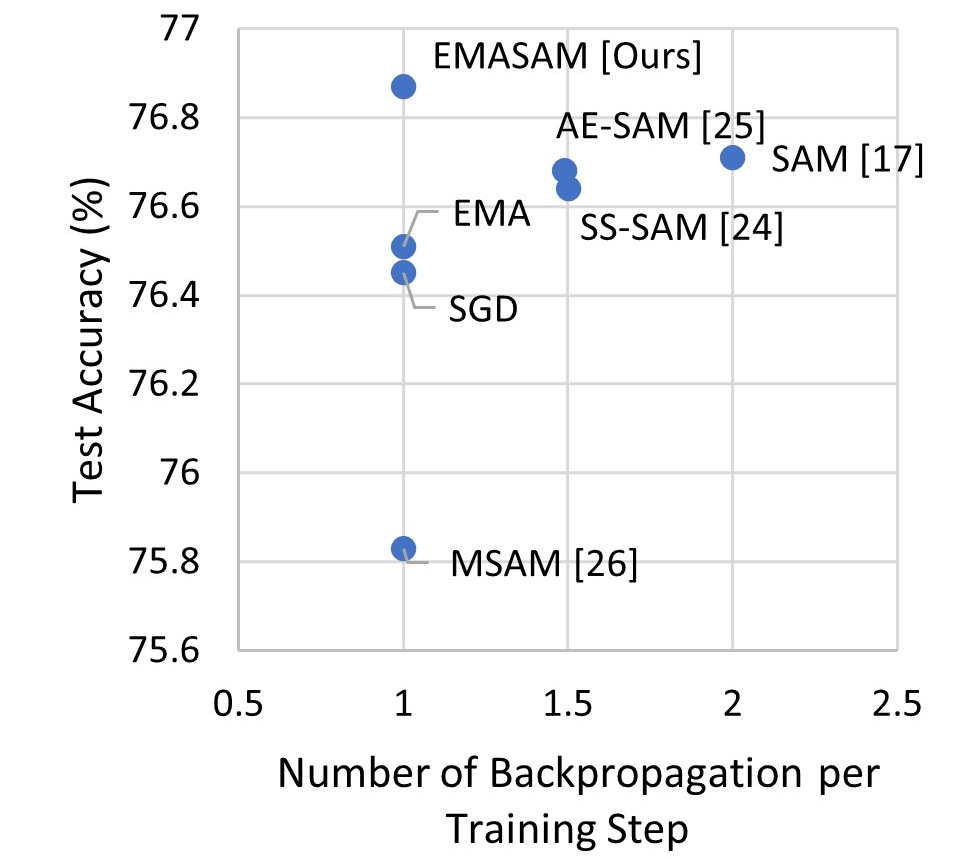}%
    }\\[1ex] % line break between rows
    
    \vspace*{-2mm}
    \caption{Number of backpropagation per training step and test accuracy}
    \label{fig:2}
    \vspace*{-5mm}
\end{figure*}

\subsection{Analysis on training overheads and model performances}
\vspace*{-2mm}
To evaluate the efficiency of EMASAM compared to other methods, we examined the number of required backpropagation steps and test accuracy across methods.

As presented in Fig.~\ref{fig:2}, EMASAM achieved the best test accuracy while only requiring a single backpropagation, demonstrating the higher efficiency of the training without sacrificing the robustness. Note that the result of the EMA with random perturbation is omitted from Fig.~\ref{fig:2}~(b) because it is out of range.

\section{Conclusion}
\label{sec:conclusion}
In this paper, we present Exponential Moving Average Sharpness-Aware Minimization (EMASAM), an efficient alternative to SAM that reduces computational overhead. Unlike SAM, EMASAM does not rely on loss-gradient information during the perturbation step. Instead, it derives the perturbation direction from the discrepancy between the current model parameters and their EMA-based shadow. By pushing the model away from a temporally stable reference toward a less stable configuration, this perturbation serves as a lightweight approximation to the worst-case direction. Furthermore, because EMASAM determines its perturbation independently of stochastic mini-batch gradients, it alleviates the instability introduced by gradient noise that commonly affects SAM. As a result, EMASAM removes the additional backpropagation required by SAM while retaining its generalization advantages. Extensive experiments further demonstrate the effectiveness and robustness of the proposed method.

\vspace*{-2mm}
\subsubsection{\ackname} This work was partially supported by JSPS KAKENHI Grant Numbers 24K02957 and was carried out using the TSUBAME4.0 supercomputer at Institute of Science Tokyo.
\vspace*{-2mm}

\bibliographystyle{splncs04}
\bibliography{ref}

\end{document}